\documentclass{article}

\usepackage[final, nonatbib]{score_workshop}

\usepackage[utf8]{inputenc} 
\usepackage[T1]{fontenc}    
\usepackage{hyperref}       
\usepackage{url}            
\usepackage{booktabs}       
\usepackage{amsfonts}       
\usepackage{nicefrac}       
\usepackage{microtype}      
\usepackage{xcolor}         
\usepackage{amsmath,amsthm}
\usepackage{caption}
\usepackage{subcaption}
\usepackage{graphicx}
\usepackage{float}
\usepackage{algorithm}
\usepackage{algpseudocode}

\title{Proposal of a Score Based Approach to Sampling Using Monte Carlo Estimation of Score and Oracle Access to Target Density}

\author{%
  Curtis J.~ McDonald\\
  Department of Statistics and Data Science\\
  Yale University\\
  \texttt{curtis.mcdonald@yale.edu} \\
   \And
   Andrew Barron \\
   Department of Statistics and Data Science \\
   Yale University\\
  \texttt{andrew.barron@yale.edu} \\
}

\begin{document}

\maketitle

\begin{abstract}
 Score based approaches to sampling have shown much success as a generative algorithm to produce new samples from a target density given a pool of initial samples. In this work, we consider if we have no initial samples from the target density, but rather $0^{th}$ and $1^{st}$ order oracle access to the log likelihood. Such problems may arise in Bayesian posterior sampling, or in approximate minimization of non-convex functions. Using this knowledge alone, we propose a Monte Carlo method to estimate the score empirically as a particular expectation of a random variable. Using this estimator, we can then run a discrete version of the backward flow SDE to produce samples from the target density. This approach has the benefit of not relying on a pool of initial samples from the target density, and it does not rely on a neural network or other black box model to estimate the score.
\end{abstract}

\section{Introduction}
\let\thefootnote\relax\footnotetext{Code available at https://github.com/CMcDonald-1/Score\_Modeling\_Monte\_Carlo}

Producing samples from a known probability measure is a core problem in statistics and many machine learning applications. Computing complicated integrals, determining the volume of a high dimensional set, or performing Bayesian inference on a posterior are all problems which require generating samples from a probability measure in question. In this work, we consider sampling from a probability measure $P_{0}$ over $\theta \in \mathbb{R}^{d}$ absolutely continuous with respect to Lebesgue measure and admitting a probability density proportional to $p_{0}(\theta) \propto e^{f(\theta) - \frac{1}{2}\theta^{T}\theta}$ for some function $f(\theta)$. The form of the density is known and we are able to query $f(\theta)$ and $\nabla f(\theta)$ at will. 

When $f(\theta)$ is concave, such a density is called log concave and can easily be sampled by known methods such as Langevin diffusion or other classical Markov Chain Monte Carlo (MCMC) methods. However, we are particularly interested in when $f(\theta)$ is non-concave having multiple separated maximizers. Traditional MCMC methods for such a density have no guarantee of convergence to invariance in finite time. Two examples are if $f(\theta)$ represents the log likelihood of a mixture model (e.g. Gaussian mixture model) and $p_{0}(\theta)$ is the posterior under a normal prior. Under such a model, the posterior has multiple modes of equal likelihood which are difficult for traditional methods to sample. If $f(\theta)$ represents a non-concave objective we want to maximize under a $L^{2}$ penalty, $\theta^{*}= \text{argmax}_{\theta} f(\theta) - \frac{1}{2}\theta^{T}\theta$, we may settle not for finding $\theta^{*}$, but rather sampling $\theta \sim p_{0}(\theta)$ and anticipating that $f(\theta^{*}) \approx f(\theta)$, that is our sampled point, while not the global maximizer, produces a function value close to the optimal value.

Thus, there remains non-log concave sampling problems yet to be sufficiently addressed by existing MCMC algorithms. In recent years, score based methods have shown remarkable success as a generative model on complex multi-modal densities in high dimensions. In this paper, we would like to transfer this success as a generative model into a successful algorithm for sampling a known density function. The main difference between a generative modelling problem and the sampling problem studied here is what ``information''  about the target density $p_{0}(\theta)$ the algorithm may access. In the current applications of score based methods as a generative algorithm, one starts with an initial pool of samples drawn from the target measure, $\theta_{0}, \cdots, \theta_{N} \sim P_{0}$, but one does not know the functional form of the target density nor can one query $f(\theta)$ and it's gradient $\nabla f(\theta)$ directly. In this work, we have no initial samples from the target density, but do have direct axis to the density function $p_{0} = e^{f(\theta) - \frac{1}{2}\theta^{T}\theta}$ as well as access to $f(\theta)$ and $\nabla f(\theta)$.

What connects both problems is the core theory of score based sampling. Underlying all score based approaches to sampling is the dual relationship between the ``forward flow'' and ``backward flow'' stochastic differential equations (SDE). Consider the forward flow SDE
\begin{align}
d\theta_{t}&= d(\theta_{t}, t)dt+g(t)dW_{t}, \theta_{0} \sim P_{0},
\end{align}
where the marginal distribution of $\theta_{t}$ at time $t$ is denoted $P_{t}$ with density $p_{t}$. Paired with this forward flow SDE is a backward flow SDE \cite{anderson1982reverse}
\begin{align}
d\theta_{t}&= [d(\theta_{t}, t)-g(t)^{2}\nabla \log p_{t}(\theta_{t})]dt+g(t)dW_{t},
\end{align}
where we run time in reverse from $t = T$ to $t = 0$. The forward flow SDE takes $\theta_{0} \sim P_{0}$ to $\theta_{T} \sim P_{T}$, while the backward flow SDE takes $\theta_{T} \sim P_{T}$ to $\theta_{0} \sim P_{0}$. If we can sample $P_{T}$ directly and implement the backward flow SDE then we can produce samples from $P_{0}$.

Crucial to the backward flow SDE is the score function, $\nabla \log p_{t}(\theta)$, which is not known and must be estimated. Popular generative approaches have a large pool of initial samples drawn from $P_{0}$. They then train a model such as a neural network to approximate the score at different time scales $t$ by minimizing an objective function using the initial samples and noisy perturbations as training data \cite{song2019generative, song2020score}. Various related methods such as \cite{de2021diffusion}, \cite{dockhorn2021critically} propose improved methods to estimate the score, but are also ultimately generative and thus rely on initial samples. \cite{doucet2022score} applies score based diffusions as an approach to improve annealed importance sampling. It is not a generative approach as it does not rely on initial samples from the target density, but it ultimately trains a neural network to estimate the score by minimizing relative entropy. In this paper, we desire a method to estimate the score without any initial samples from $P_{0}$ and without appealing to a complex secondary model such as a neural network to estimate the score.

As a summary of what is to come, with knowledge of the functional form of the density $p_{0}$ and application of stochastic calculus, we can analytically express how $p_{t}$ changes over time and express $p_{t}$ as an expectation of a known function over a known density. With some further analysis, the score $\nabla  \log p_{t}(\theta)$ can also be expressed as the expectation of a function over a density, and thus the score can be computed by estimating this integral via a sub Monte Carlo (MC) sampling problem. This replaces the need to estimate the score via a neural network with estimating the score via a MC average.

Specifically, we will consider the forward flow SDE as the Ornstein–Uhlenbeck process with drift $d(\theta,t)=-\theta$ and constant diffusion $g(t) = \sqrt{2}$
\begin{align}
d\theta_{t}&= -\theta_{t}dt+\sqrt{2}dW_{t}.
\end{align}
The invariant measure is standard normal and the conditional measure $p_{t}(\theta_{t}|\theta_{0})$ can be expressed as
\begin{align}
\theta_{t}&= e^{-t}\theta_{0}+\sqrt{1-e^{-2t}}Z, Z \sim N(0, I).\label{OU_soln}
\end{align}
Thus the density $p_{t}(\theta_{t})$ up to normalization can be computed by integrating $p_{0}(\theta_{0})p_{t}(\theta_{t}|\theta_{0})$ over $\theta_{0}$ and the score, which removes any dependence on normalizing constants, can be expressed as
\begin{align}
\nabla_{\theta_{t}} \log p_{t}(\theta_{t}) &= \frac{\int p_{0}(\theta_{0})\nabla_{\theta_{t}}p_{t}(\theta_{t}|\theta_{0})d\theta_{0}}{\int p_{0}(\theta_{0})p_{t}(\theta_{t}|\theta_{0})d\theta_{0}}.\label{score_ratio}
\end{align}
With some manipulations, we can express the numerator and denominator as expectations over normal random variables, and thus can approximate each via averaging the integrand at normally drawn points, i.e. a MC estimator.

\section{Results}

Simplifying equation (\ref{score_ratio}) (see appendix for full derivation) we can express the score as a  ratio of two expectations,
\begin{align}
\nabla \log p_{t}(\theta)&=- \theta+\frac{e^{-t}}{\sqrt{1-e^{-2t}}}\frac{E[Ue^{f(\sqrt{1-e^{-2t}}U+e^{-t}\theta)}]}{E[e^{f(\sqrt{1-e^{-2t}}U+e^{-t}\theta)}]}\quad U \sim N(0,I),\label{expression_1}
\end{align}
applying integration by parts (see appendix), this expression can equivalently be written as
\begin{align}
\nabla \log p_{t}(\theta)&= -\theta+e^{-t}\frac{E[\nabla f(\sqrt{1-e^{-2t}}U+e^{-t}\theta)e^{f(\sqrt{1-e^{-2t}}U+e^{-t}\theta)}]}{E[e^{f(\sqrt{1-e^{-2t}}U+e^{-t}\theta)}]}\quad U \sim N(0, I).\label{expression_2}
\end{align}

Computing expressions (\ref{expression_1}) and (\ref{expression_2}) then amounts to computing expectations over normal random variables, and we will appeal to a MC estimator. Say we draw $K$ points $U_{1}, \cdots, U_{K}\sim N(0,I)$ from a standard normal, and approximate the score in one of two ways. Define weights $w_{k}$ which sum to 1
\begin{align}
w_{k}&= \frac{e^{f(\sqrt{1-e^{-2t}}U_{k}+e^{-t}\theta)}}{\sum_{j=1}^{K} e^{f(\sqrt{1-e^{-2t}}U_{j}+e^{-t}\theta)}},
\end{align}
as the relative heights of the exponential of $f$ evaluated at the sampled points. We can then approximate (\ref{expression_1}) and (\ref{expression_2}) as weighted averages using these weightings
\begin{align}
\hat{s}_{1}(\theta, t)&= -\theta + \frac{e^{-t}}{\sqrt{1-e^{-2t}}} \sum_{k=1}^{K}U_{k} w_{k}\\
\hat{s}_{2}(\theta, t)&= -\theta+e^{-t} \sum_{k=1}^{K} \nabla f(\sqrt{1-e^{-2t}}U_{k}+e^{-t}\theta)w_{k}.\label{s_2}
\end{align}
There are two perspectives on how to estimate the ratio of expectations that define (\ref{expression_1}) and (\ref{expression_2}). One could use a separate pool of samples to evaluate the numerator and denominator expectations and then take the ratio of these two estimates. However, we may be dividing by a potentially small number in this method and have high variance as a result. The authors here propose using the same pool of samples to estimate both the numerator and denominator, resulting in the weights $w_{k}$ where the normalizing constant of the weights represents the denominator in (\ref{expression_1}) and (\ref{expression_2}). This has the advantage that the estimates will never become unbounded since all estimates are weighted averages of well behaved values, however proper analysis must be conducted on how such a sampling procedure effects the independence, bias, and variance of the estimators.

For small $t$, say $t< 0.1$, we conjecture that $\hat{s}_{2}(\theta, t)$ is the better estimator as it will have lower variance since $\frac{e^{-t}}{\sqrt{1-e^{-2t}}}$ can grow large as $t \to 0$. Additionally, as $t \to 0$, $\nabla f(\sqrt{1-e^{-2t}}U+e^{-t}\theta) \to \nabla f(\theta)$ and estimate $\hat{s}_{2}(\theta, t)$ approaches $-\theta+\nabla f(\theta)$ for small $t$, that is the score of the target density.

For larger $t$, it would seem $\hat{s}_{1}(\theta, t)$ will have lower variance since $U_{k}$ is less variable than the gradient of the target $\nabla f(U_{k})$. Additionally, as $t \to \infty$ we have the weights $w_{k}$ approach a $\theta$ independent value due to the $e^{-t}\theta$,  so the expectation is some finite vector while $\frac{e^{-t}}{\sqrt{1-e^{-2t}}} \to 0$ thus the score approaches $-\theta$, the score of a standard normal.

 The algorithm to generate a sample from $P_{0}$ is as follows. Pick a terminal time $T>0$, step size $\delta>0$, and MC sample size $K \in \mathbb{N}$. Initialize $\theta_{T} \sim N(0,I)$ and $t = T$. While $t>0$, in each iteration draw $K$ values $U_{1},\cdots, U_{K} \sim N(0,I)$ and compute weights $w_{k}$. For large times $t>0.1$ estimate score as $\hat{s}(\theta_{t}, t)=\hat{s}_{1}(\theta, t)$, otherwise set $\hat{s}(\theta_{t}, t)=\hat{s}_{2}(\theta, t)$. Update the point as $
\theta_{t-\delta}=\theta_{t}-\delta(-\theta_{t}-2\hat{s}(\theta_{t}, t))+\sqrt{2\delta}Z,\quad Z \sim N(0, I)$ and decrease $t = t-\delta$. Finally, return $\theta_{0}$.

One could propose more complex methods such Stein's method \cite{chen2019stein},\cite{liu2016stein} or multi-modal sampling methods such as Simulated Tempering \cite{ge2018simulated} to estimate the quantities in (\ref{expression_1}) and (\ref{expression_2}), as these represent expectations over un-normalized densities (the denominator is the normalizing constant). However, estimating these values is itself a sub problem of sampling the un-normalized density $p_{0}$ in the first place, thus if we employ sufficiently complex methods to sample the sub problems we may have simply employed said methods on the original problem. Additionally, for large $t>>0$ and small $t \approx 0$ the simple estimators will have small variance, thus it is the middle region of $t$ where it may require more involved estimation of quantities (\ref{expression_1}) and (\ref{expression_2}) to keep the estimates accurate.

\section{Example}

Here we demonstrate the performance of the algorithm on a series of example problems. These examples while low dimensional represent multi-modal targets with future work considering higher dimensional implementations. First, we sample a one dimensional multi-modal target density with asymmetric modes. Using the log likelihood function
\begin{align*}
f(\theta)&= 100\sum_{i=1}^{4}(\tanh(\theta+0.05-\mu_{i})-\tanh(\theta-0.05-\mu_{i})),\mu_{1}=-5, \mu_{2}=-1, \mu_{3}=3, \mu_{4}=4.
\end{align*}

\begin{figure}
\begin{subfigure}[b]{0.5\textwidth}
      \centering
        \includegraphics[width = \textwidth]{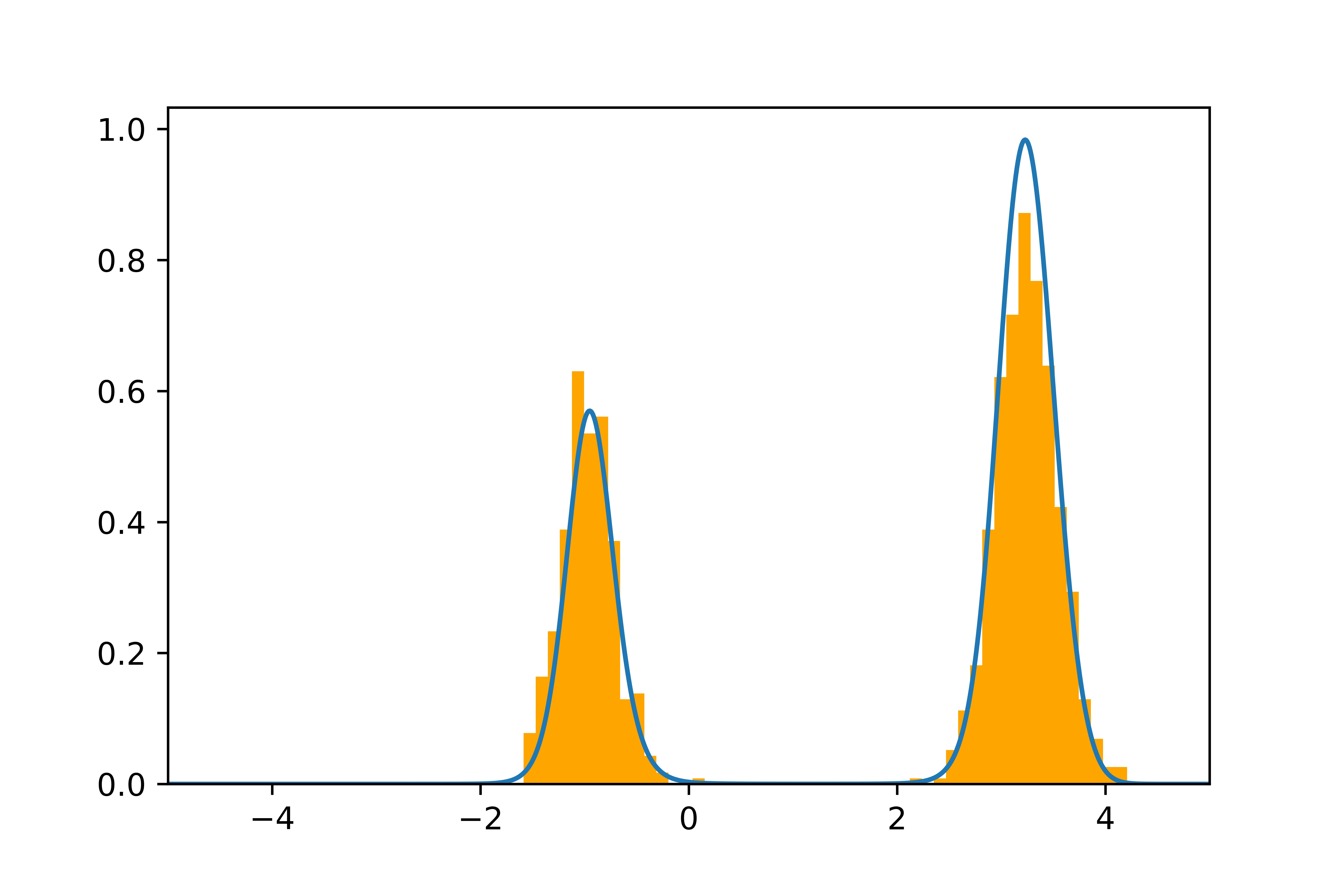}
        \caption{One Dimensional Sampling Histogram}
         \label{fig:one_dim}
\end{subfigure}
     \hfill
    \begin{subfigure}[b]{0.5\textwidth}
        \centering
        \includegraphics[width = \textwidth]{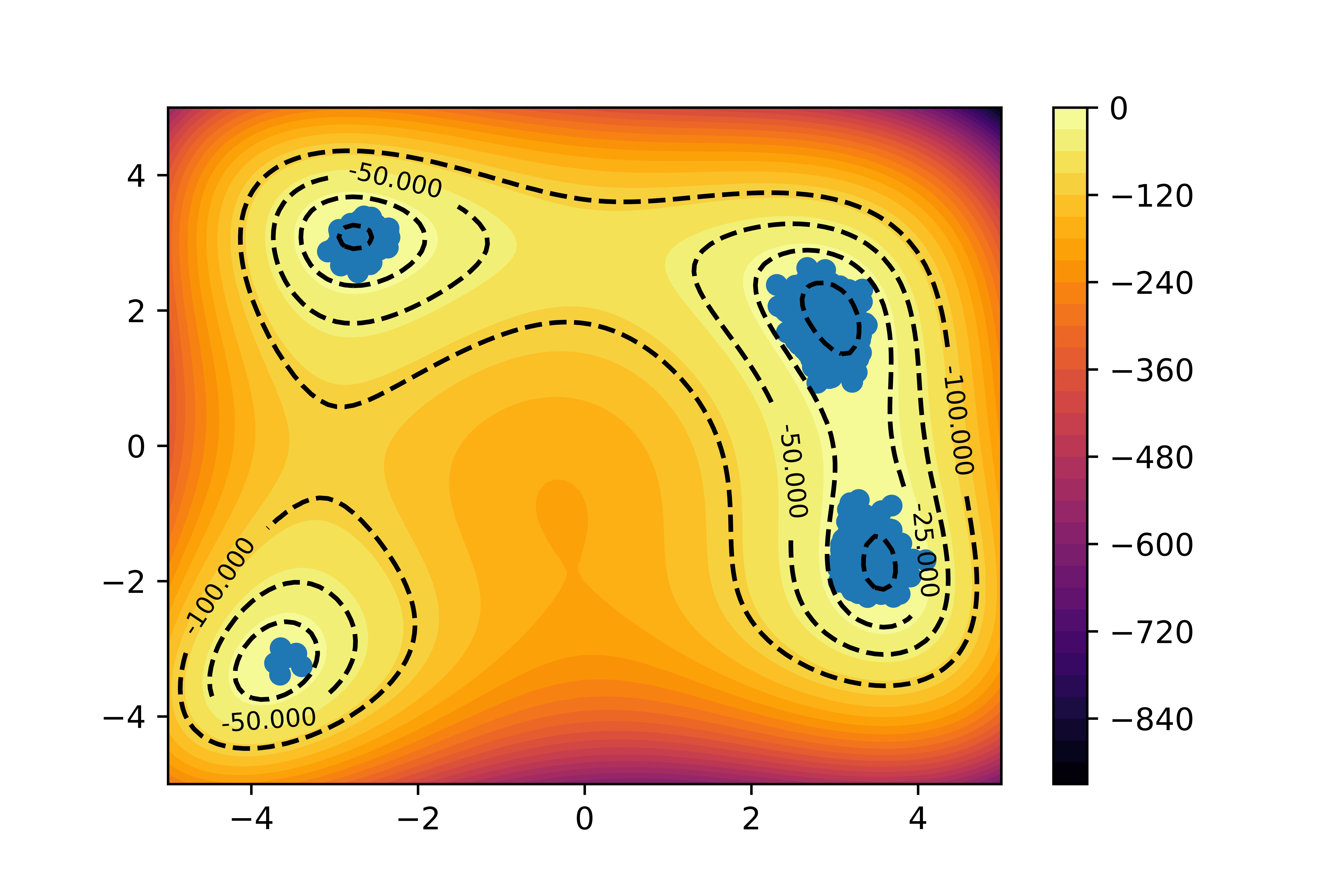}
         \caption{Himmelblau Sampled Points}
         \label{fig:himmelblau}
\end{subfigure}
\caption{Samples produced for 1 and 2 dimensional sampling problems.}
\end{figure}

In Figure \ref{fig:one_dim} we see the histogram of $n = 1000$ points ran in parallel with terminal time $T = 2$, $K = 3000$ samples per estimate, and step size $\delta = 0.01$. This requires $K*(T/\delta) = 600,000$ evaluations of $f$ and $\nabla f$ per point, a significantly high number of evaluations per sample but the polynomial nature may see better gains in high dimensions compared with existing methods.

Another typical problem to test multi-modal sampling is to sample the Himmelblau function	\cite{feroz2013exploring} in two dimensions
\begin{align*}
f(\theta_{1}, \theta_{2})&= -(\theta_{1}^{2}+\theta_{2}-11)^{2}-(\theta_{1}+\theta_{2}^{2}-7)^{2}
\end{align*}
which has modes at $(3,2), (-2.81, 3.13), (-3.78, -3.28), (3.58, -1.85)$ which are all separated by areas of low probability and asymmetric in their relative heights. Sampling $n = 2000$ points in parallel with $T = 3$, $K = 1000$, $\delta = 0.01$ this algorithm requires $K*(T/\delta) = 300,000$ evaluations of $f$ and $\nabla f$ per point. To test if each mode is sampled proportionally correct, we compute the proportion of points within $\pm 0.5$ in each coordinate around each mode and normalize to sum to 1. We then evaluate the target pdf at $10,000$ points sampled uniformly around each mode as an estimate of the probability of the region around each mode and normalize to sum to 1. If each more is sampled proportionally correctly, then these two vectors should be the same. The probability estimates of each mode's immediate vicinity are (0.831, 0.009, 0.002, 0.158) while the proportion of points near each mode is (0.741, 0.053, 0.005, 0.201). We note that the modes are not sampled exactly in the correct proportions, but the proportions are reasonable close to each mode, and each mode is visited, so as an approximate sampling algorithm this may suffice in certain situations.

\section{Concluding remarks}
This paper presents a score based sampling approach using Monte Carlo estimation of the score and knowledge of the functional form of the target density. This method does not rely on any initial samples from the target density, nor on a model such as a neural network to estimate the score function. Much work remains to be done on the the analysis of such a method, specifically on high dimensional target densities of interest. The MC estimators may be of high variance, and theoretical guarantees on this variance must be provided. Furthermore, the estimates of the score will never be perfect, they will have some standard error around the true value, thus analysis of the backward flow SDE under discretization and approximate drift such as the analysis in \cite{lee2022convergence} will be also be required to guarantee convergence to the correct target measure.

\bibliographystyle{plain}
\bibliography{my_score_flow_MC_paper_v2}

\section{Appendix}
Here we derive the form of the expectations that define the score. Define $\sigma_{t} = \sqrt{1-e^{-2t}}$, we have
\begin{align*}
\nabla_{\theta_{t}} \log p_{t}(\theta_{t})&= \frac{\int p_{0}(\theta_{0})\nabla_{\theta_{t}}p_{t}(\theta_{t}|\theta_{0})d\theta_{0}}{\int p_{0}(\theta_{0})p_{t}(\theta_{t}|\theta_{0})d\theta_{0}}\\
&= \frac{\int e^{f(\theta_{0})-\frac{1}{2}\theta_{0}^{T}\theta_{0}}\nabla_{\theta_{t}}e^{-\frac{1}{2 \sigma_{t}^{2}}(\theta_{t}-e^{-t}\theta_{0})^{T}(\theta_{t}-e^{-t}\theta_{0})}d\theta_{0}}{\int e^{f(\theta_{0})-\frac{1}{2}\theta_{0}^{T}\theta_{0}}e^{-\frac{1}{2 \sigma_{t}^{2}}(\theta_{t}-e^{-t}\theta_{0})^{T}(\theta_{t}-e^{-t}\theta_{0})}d\theta_{0}}\\
&=\frac{\int e^{f(\theta_{0})}\left(- \frac{\theta_{t}-e^{-t}\theta_{0}}{\sigma_{t}^{2}}\right)e^{-\frac{1}{2}\theta_{0}^{T}\theta_{0}-\frac{1}{2\sigma_{t}^{2}}(\theta_{t}-e^{-t}\theta_{0})^{T}(\theta_{t}-e^{-t}\theta_{0})}d\theta_{0}}{\int e^{f(\theta_{0})}e^{-\frac{1}{2}\theta_{0}^{T}\theta_{0}-\frac{1}{2\sigma_{t}^{2}}(\theta_{t}-e^{-t}\theta_{0})^{T}(\theta_{t}-e^{-t}\theta_{0})}d\theta_{0}}
\end{align*}
we now perform some completing the square on the $\theta_{0}$ terms in the exponent to express it as a normal density over $\theta_{0}$. Note by cancellation on the top and bottom of the fraction, we can ignore any constant terms in the integral and only focus on those terms which relate to $\theta_{0}$.
\begin{align*}
&e^{f(\theta_{0})-\frac{1}{2}\theta_{0}^{T}\theta_{0}}e^{-\frac{1}{2\sigma_{t}^{2}}(\theta_{t}-e^{-t}\theta_{0})^{T}(\theta_{t}-e^{-t}\theta_{0})}\\
&= e^{f(\theta_{0})}e^{-\frac{1}{2}[\theta_{0}^{T}\theta_{0}+ \frac{1}{\sigma_{t}^{2}}(\theta_{t}^{T}\theta_{t}-2e^{-t}\theta_{t}^{T}\theta_{0}+e^{-2t}\theta_{0}^{T}\theta_{0})]}\\
&\propto e^{f(\theta_{0})}e^{-\frac{1}{2} ((1+\frac{e^{-2t}}{1-e^{-2t}})\theta_{0}^{T}\theta_{0}-2\frac{e^{-t}}{1-e^{2t}}\theta_{t}^{T}\theta_{0})}\\
&\propto e^{f(\theta_{0})}e^{-\frac{1}{2\sigma_{t}^{2}}(\theta_{0}-e^{-t}\theta_{t})^{T}(\theta_{0}-e^{-t}\theta_{t})}
\end{align*}
thus, the above function is proportional to $e^{f(\theta_{0})}N(e^{-t}\theta_{t}, \sigma_{t}^{2}I)$ and we can write the score as
\begin{align*}
\frac{- E[e^{f(\theta_{0})}\frac{\theta_{t}-e^{-t}\theta_{0}}{\sigma_{t}^{2}}]}{E[e^{f(\theta_{0})}]}\quad \theta_{0}\sim N(e^{-t}\theta_{t}, \sigma_{t}^{2}I)
\end{align*}
then, consider if instead $U \sim N(0, I)$ and $\theta_{0}  = \sigma_{t}U+e^{-t}\theta_{t}$, we could write the above as
\begin{align*}
&\frac{-E[e^{f(\sigma_{t}U+e^{-t}\theta_{t})} \frac{\theta_{t}-e^{-t}\sigma_{t}U-e^{-2t}\theta_{t}}{1-e^{-2t}}]}{E[e^{f(\sigma_{t}U+e^{-t}\theta_{t})}]}\\
&= \frac{-E[e^{f(\sigma_{t}U+e^{-t}\theta_{t})}(\theta_{t}-\frac{e^{-t}}{\sigma_{t}}U)]}{E[e^{f(\sigma_{t}U+e^{-t}\theta_{t})}]}\\
&= -\theta_{t}+ \frac{e^{-t}}{\sigma_{t}}\frac{E[Ue^{f(\sigma_{t}U+e^{-t}\theta_{t})}]}{E[e^{f(\sigma_{t}U+e^{-t}\theta_{t})}]}
\end{align*}
We can also apply integration by parts in the numerator to arrive at an equivalent expression,
\begin{align*}
\frac{e^{-t}}{\sigma_{t}}E[Ue^{f(\sigma_{t}U+e^{-t}\theta_{t})}]&=\frac{e^{-t}}{\sigma_{t}}\int ue^{f(\sigma_{t}u+e^{-t}\theta_{t})}\left(\frac{1}{2\pi}\right)^{\frac{d}{2}}e^{-\frac{1}{2}u^{T}u}du\\
&=-\frac{e^{-t}}{\sigma_{t}}\int e^{f(\sigma_{t}u+e^{-t}\theta_{t})}\left(-u\left(\frac{1}{2\pi}\right)^{\frac{d}{2}}e^{-\frac{1}{2}u^{T}u}\right)du\\
&=-\frac{e^{-t}}{\sigma_{t}}\int e^{f(\sigma_{t}u+e^{-t}\theta_{t})}\nabla_{u}\left(\left(\frac{1}{2\pi}\right)^{\frac{d}{2}}e^{-\frac{1}{2}u^{T}u}\right)du\\
&=\frac{e^{-t}}{\sigma_{t}}\int \nabla_{u}\left(e^{f(\sigma_{t}u+e^{-t}\theta_{t})}\right)\left(\frac{1}{2\pi}\right)^{\frac{d}{2}}e^{-\frac{1}{2}u^{T}u}du\\
&=\frac{e^{-t}}{\sigma_{t}}\int\sigma_{t}\nabla f(\sigma_{t}u+e^{-t}\theta_{t})e^{f(\sigma_{t}u+e^{-t}\theta_{t})}\left(\frac{1}{2\pi}\right)^{\frac{d}{2}}e^{-\frac{1}{2}u^{T}u}du\\
&=e^{-t}E[\nabla f(\sigma_{t}U+e^{-t}\theta_{t})e^{f(\sigma_{t}U+e^{-t}\theta_{t})}]
\end{align*}
and thus can express the score as
\begin{align*}
-\theta_{t}+ e^{-t}\frac{E[\nabla f(\sigma_{t}U+e^{-t}\theta_{t})e^{f(\sigma_{t}U+e^{-t}\theta_{t})}]}{E[e^{f(\sigma_{t}U+e^{-t}\theta_{t})}]}
\end{align*}

\end{document}